\documentclass[11pt,a4paper]{article}
\usepackage[hyperref]{eacl2021}
\usepackage{times}
\usepackage{latexsym}

\usepackage{microtype}

\aclfinalcopy 
\def\aclpaperid{47} 

\author{Fatima Haouari, Maram Hasanain, Reem Suwaileh, Tamer Elsayed \\
  Computer Science and Engineering Department, Qatar University \\
  \texttt{\{200159617, maram.hasanain, rs081123, telsayed\}@qu.edu.qa}}

\date{}

\usepackage{array}
\usepackage{soul}
\usepackage{xcolor}
\usepackage{graphicx}
\graphicspath{ {./figures/} }

\usepackage{tabularx}
\usepackage{enumitem}
\usepackage{booktabs}
\usepackage{makecell}

\definecolor{lightblue}{rgb}{.50,.95,1}
\definecolor{tri}{rgb}{.25,.88,.82}
\definecolor{lilac}{rgb}{0.85,0.64,0.85}

\newcommand{\ds}{{ArCOV-19}}
\title{\ds: The First Arabic COVID-19 Twitter Dataset \\with Propagation Networks}
\begin{document}
\maketitle
\begin{abstract}
In this paper, we present \ds{}, an Arabic COVID-19 Twitter dataset that spans one year, covering the period from 27$^{th}$ of January 2020 till 31$^{st}$ of January 2021. \ds{} is the \emph{first} publicly-available Arabic Twitter dataset covering COVID-19 pandemic that includes about 2.7M tweets 
alongside the \emph{propagation networks} of the most-popular subset of them (i.e., most-retweeted and -liked). The propagation networks include both retweets and conversational threads (i.e., threads of replies). \ds{} is designed to enable research under several domains including natural language processing, information retrieval, and social computing
. Preliminary analysis shows that \ds{} captures rising discussions associated with the first reported cases of the disease as they appeared in the Arab world. In addition to the source tweets and propagation networks, we also release the search queries and language-independent crawler used to collect the tweets to encourage the curation of similar datasets. 
\end{abstract}
\section{Introduction}
\label{intro}

Twitter streams hundreds of millions of tweets daily. In addition to being a medium for the spread and consumption of news, it has been shown to capture the dynamics of real-world events including the spread of diseases such as the seasonal influenza~\cite{kagashe2017} or more severe epidemics like Zika~\cite{vijaykumar2018virtual}, Ebola~\cite{roy2020ebola}, and H1N1~\cite{McNeill2016}. Moreover, collective conversations on Twitter about an event can have a great influence on the event's outcomes, e.g., US 2016 presidential elections~\cite{GROVER2019438}. 
Analyzing tweets about an event, as it evolves, offers a great opportunity to understanding its structure and characteristics, informing decisions based on its development, and anticipating its outcomes as represented in Twitter and, more importantly, in the real world. 

Since the first reported case of Novel Coronavirus (COVID-19) in China, in November 2019, the COVID-19 topic has drawn the interest of many Arab users over Twitter. Their interest, reflected in the Arabic content on the platform, has reached a peak after two months when the first case was reported in the United Arab Emirates late in January 2020. This ongoing pandemic has, unsurprisingly, spiked discussions on Twitter covering a wide range of topics such as general information about the disease, preventive measures, procedures and newly-enforced decisions by governments, up-to-date statistics of the spread in the world, and even the change in our daily habits and work styles.

In this work, we aim to support future research on social media during this historical period of our time by curating an Arabic dataset (\ds) that exclusively covers tweets about COVID-19. We limit the dataset to Arabic since it is among the most dominant languages in Twitter~\cite{alshaabi2020}, yet under-studied in general. 

\ds{} is the \emph{first} Arabic Twitter dataset designed to capture tweets discussing COVID-19 starting from January $27^{th}$ 2020 till end of January 2021\footnote{The dataset will be continuously augmented with new tweets over the coming months.}, constituting about 2.7M tweets, alongside the propagation networks of the most-popular subset of them. To our knowledge, there is no publicly-available Arabic Twitter dataset for COVID-19 that includes the propagation networks of a good subset of its tweets. Some existing efforts have already started to curate COVID-19 datasets including Arabic tweets, but Arabic is severely under-represented (e.g.,~\cite{chen2020covid,singh2020first}) or represented by a random sample that is not specifically focused on COVID-19 (e.g.,~\cite{alshaabi2020world}), or the dataset is limited in the period it covers and does not include the propagation networks (e.g.,~\cite{alqurashi2020large}).

The contribution of this paper is three-fold:
\begin{itemize}[noitemsep]

      \item We release \ds,\footnote{\url{https://gitlab.com/bigirqu/ArCOV-19/}} the first Arabic Twitter dataset about COVID-19 that comprises about 2.7M tweets collected via Twitter search API. It covers a full year allowing for capturing discussions on the topic since it started to be popular in the Arab world. 
    In addition to the tweets, \ds{} includes propagation networks of the most popular subset, search queries,
    and documented implementation of our language-independent tweets crawler.
    \item We present a preliminary analysis on \ds{}, which reveals insights from the dataset concerning temporal, geographical, and topical aspects. 
    \item We suggest several use cases to enable research on Arabic tweets in different research areas including, but not limited to, emergency management, misinformation detection, and social analytics.
\end{itemize}

The reminder of the paper is organized as follows. We present the crawling process followed to acquire \ds{} in Section~\ref{data}. We then thoroughly discuss the analysis that we conducted on the dataset in Section~\ref{discuss}. We suggest some use cases to enable research on Arabic tweets in Section~\ref{usecases}. We finally conclude in Section~\ref{conc}.
\section{Related Work}

Social media platforms and Twitter specifically showed to be an indispensable medium for sharing information and discussions about the COVID-19 pandemic since December 2019.
A considerable body of raw social media datasets were released to facilitate analyzing these discussions including Twitter datasets~\cite{banda2020large,lopez2020understanding,chen2020tracking,qazi2020geocov19,gao2020naist,alqurashi2020large,dashtian2021cml}, Instagram~\cite{zarei2020first}, or the Chinese social media platform Weibo~\cite{hu2020weibo,gao2020naist}.

 \citet{banda2020large} released a multilingual dataset that includes over 800M tweets, the data is collected since 1$^{st}$ of January 2020 using Twitter streaming API. \citet{dashtian2021cml}, collected multilingual tweets posted between March and July 2020. The most dominant languages covered by their dataset are English and Spanish constituting 65\% and 12\% of the tweets respectively.  
 \citet{chen2020tracking} are continously collecting tweets since 21$^{st}$ of January 2020, where at the time of writing, around 66\% and 11\% of the tweets are in English and Spanish respectively. Another multilingual ongoing collection is released by~\citet{lopez2020understanding}. \citet{qazi2020geocov19} enriched their large-scale multilingual Twitter dataset with geolocation information. 

The only raw Arabic Twitter dataset available is the one released by~\citet{alqurashi2020large}, however it covers the period from 1$^{st}$ of January to 15$^{th}$ April 2020 only. Moreover, it does not include the propagation networks of tweets.

Compared to existing datasets, we release an only-Arabic collection of tweets, where we exclude retweets to avoid having copies of the same tweet. Moreover, we also release the most-popular subset of them (i.e., most-retweeted and -liked and spam free tweets). Furthermore, we collect the propagation  networks including both retweets and conversational threads (i.e., threads of replies) for the most-popular subset.

\section{Data Collection}

\label{data}
\ds{} includes two major components: the \textbf{source tweets} (i.e., tweets collected via Twitter search API every day) and the \textbf{propagation networks} (i.e., retweets and conversational threads of a subset of the source tweets). In this section, we present how we collected each in Sections~\ref{tweetCollection}, and~\ref{propagationCollection} respectively, and summarize the released data in Section~\ref{sec:release}. 

\subsection{Tweets Collection}\label{tweetCollection}
 To collect the source tweets, we used our tweet crawler\footnote{\url{https://gitlab.com/bigirqu/ArCOV-19/-/tree/master/code/crawler}} that uses Twitter \emph{search} API\footnote{\url{https://developer.twitter.com/en/docs/tweets/search/api-reference/get-search-tweets}}. The crawler takes a set of manually-crafted queries, comprising a target topic, as input. At the end of each day, the crawler issues a search request for each of those queries. Queries can be 
keywords (e.g., ``Corona''), 
phrases (e.g., ``the killing virus''), 
or hashtags (e.g., ``\#the\_new\_coronavirus'').
Twitter returns a maximum of 3,200 tweets per query. We customized the search requests to return only \emph{Arabic} tweets,\footnote{Language of tweets to return is a configurable parameter.} and to exclude all retweets to avoid having copies of the same tweet. Additionally, duplicate tweets (returned by different queries) are removed in each day. Finally, tweets are sorted chronologically. 

We started collecting our data since $27^{th}$ of January 2020 using a set of queries that we manually-updated based on our daily tracking of trending keywords and hashtags. For example,  starting from $16^{th}$ of December 2020, we extended our queries to cover tweets related to COVID-19 vaccines as that sub-topic was gaining interest on Arabic social media. The full list of queries used in each day is released alongside our dataset. We denote all collected tweets as \textbf{\emph{source tweets}}.

Due to technical reasons, we missed collecting tweets for a few days. Since Twitter search API limits the search results to the past 7 days, we missed the old tweets. To overcome that, we used GetOldTweets3\footnote{\url{https://github.com/Mottl/GetOldTweets3}} python library to download the search results using the same trending keywords and hashtags we selected around those days.

\subsection{Propagation Networks Collection}\label{propagationCollection}
In addition to the source tweets, we also collected the \textbf{\emph{propagation networks}} (i.e., retweets and conversational threads) of the top 1000 most popular (i.e., top-retweeted \& -liked) tweets each day. To our knowledge, this is the \emph{first} Arabic tweet dataset to include such propagation networks. 

Before getting the most popular tweets on any day, we started from source tweets collected in that day and applied a qualification pipeline. We first excluded tweets containing any inappropriate word (using a list of inappropriate words we constructed). Next, tweets with more than two URLs or four hashtags, or shorter than four tokens (all are potentially spam) were dropped. Additionally, duplicate tweets that have exact textual content are also dropped to avoid redundancy; only the most popular of them (according to our scoring criterion below) is kept. 
Qualified tweets are then scored by popularity defined by the sum of the tweet's retweet and favorite counts. We finally sort the qualified tweets by their scores and select the most popular 1,000 tweets. We denote the set of all such tweets over all days as the \textbf{top subset}.

For those 1K top tweets per day, we then collected all retweets and conversational threads (i.e., direct and indirect replies).\footnote{At the time of writing, we have collected the replies for tweets until end of April 2020 and we are still collecting the rest; we will make all available in the near future.} 
We collected the retweets using Pickaw,\footnote{\url{https://pickaw.com/en/twrench-becomes-pickaw}\label{pickawLabel}} a platform for organizing contests on social media, and the replies using PHEME~\cite{zubiaga2016analysing} Twitter conversation collection script.\footnote{\url{https://github.com/azubiaga/pheme-twitter-conversation-collection}}

\subsection{Data Release} 
\label{sec:release}
In summary, we release the following resources as \ds{} dataset, taking into consideration Twitter content redistribution policy:\footnote{\url{https://developer.twitter.com/en/developer-terms/agreement-and-policy}}
\setlist{nolistsep}
\begin{itemize}[noitemsep]
    \item \textbf{Source Tweets}: IDs of the tweets crawled in each day.
    \item \textbf{Search Queries}: the list of search queries, including keywords, phrases, and hashtags, used in each period to collect our source tweets.
    \item \textbf{Top Subset}: IDs of the top 1,000 most popular tweets for each day.  
    \item \textbf{Propagation Networks}: the propagation networks for the top subset which include for each tweet in the top subset:
        \begin{itemize}
            \item \textbf{Retweets}: IDs of the full retweet set.
            \item \textbf{Conversational Threads}: tweet IDs of the full reply thread (including direct and indirect replies).
        \end{itemize}
\end{itemize}
Along with the dataset, we provide 
some pointers to publicly-available crawlers that users can easily use to crawl the tweets given their IDs.
\section{\ds{} in Numbers}
\label{discuss}


\urldef{\china}\url{https://tinyurl.com/rfj5khn}
\urldef{\iran}\url{https://tinyurl.com/yykgwbox}
\urldef{\kuwait}\url{https://tinyurl.com/y4nlgz5h}
\urldef{\lebanon}\url{https://tinyurl.com/yxkg78g5}
\urldef{\twitterSaudi}\url{https://tinyurl.com/jmkttt3}

In this section, we present a statistical summary and conduct an analysis on \ds{} to shed some light on its major characteristics. 
\subsection{Tweets and Users Distribution}
Table~\ref{tab:statistics} presents an overall summary of the tweets and users statistics in \ds{}. It indicates that the total number of source tweets in the dataset is about 2.7M posted by over
690k unique users. We note that
18.66\% of the tweets were posted by verified users (who constitute only 0.81\% of the unique users). This is a relatively large percentage, showing that a good portion of the source tweets are popular. The average numbers of followers, friends, and statuses of users are also relatively large, showing that users observed in \ds{} are also popular (and possibly more influential). The table also indicates that 25.40\% of the tweets include URLs. We anticipate that this is due to the extensive spreading of news (linked from tweets) during this period. The numbers of tweets that are geotagged and geolocated are also indicated in the table; however, we defer the discussion of such types of tweets to Section~\ref{sec:geo}.

\begin{figure}[t]
\centering
 \includegraphics[scale=0.38]{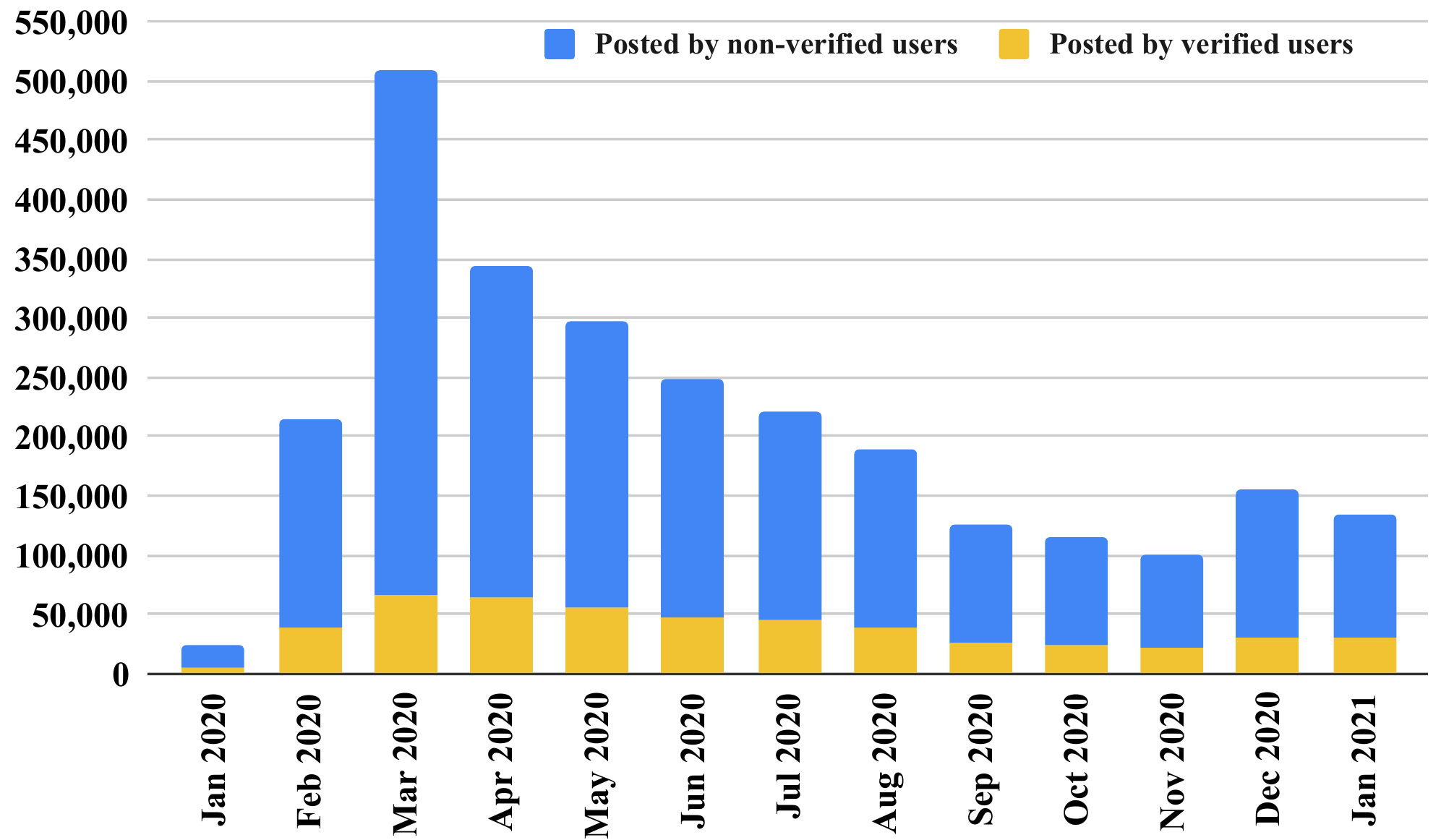}
\caption{Monthly distribution of source tweets of \ds{}. \label{fig:ntweets}}
\end{figure}

Figure~\ref{fig:ntweets} illustrates the monthly distribution of tweets in \ds. The volume of tweets drastically increased in March when the virus started to spread in several Arab countries. Then, tweets number started to decrease monthly. We speculate this is due to the fact that the topic was not heavily discussed as in the first days. In December, we expanded our queries list to include hashtags and keyword related to more current sub-topics such as vaccines and the new variant of the virus, which resulted in an increase in the volume of tweets.

\begin{figure}[h]
\centering
\includegraphics[scale=0.45]{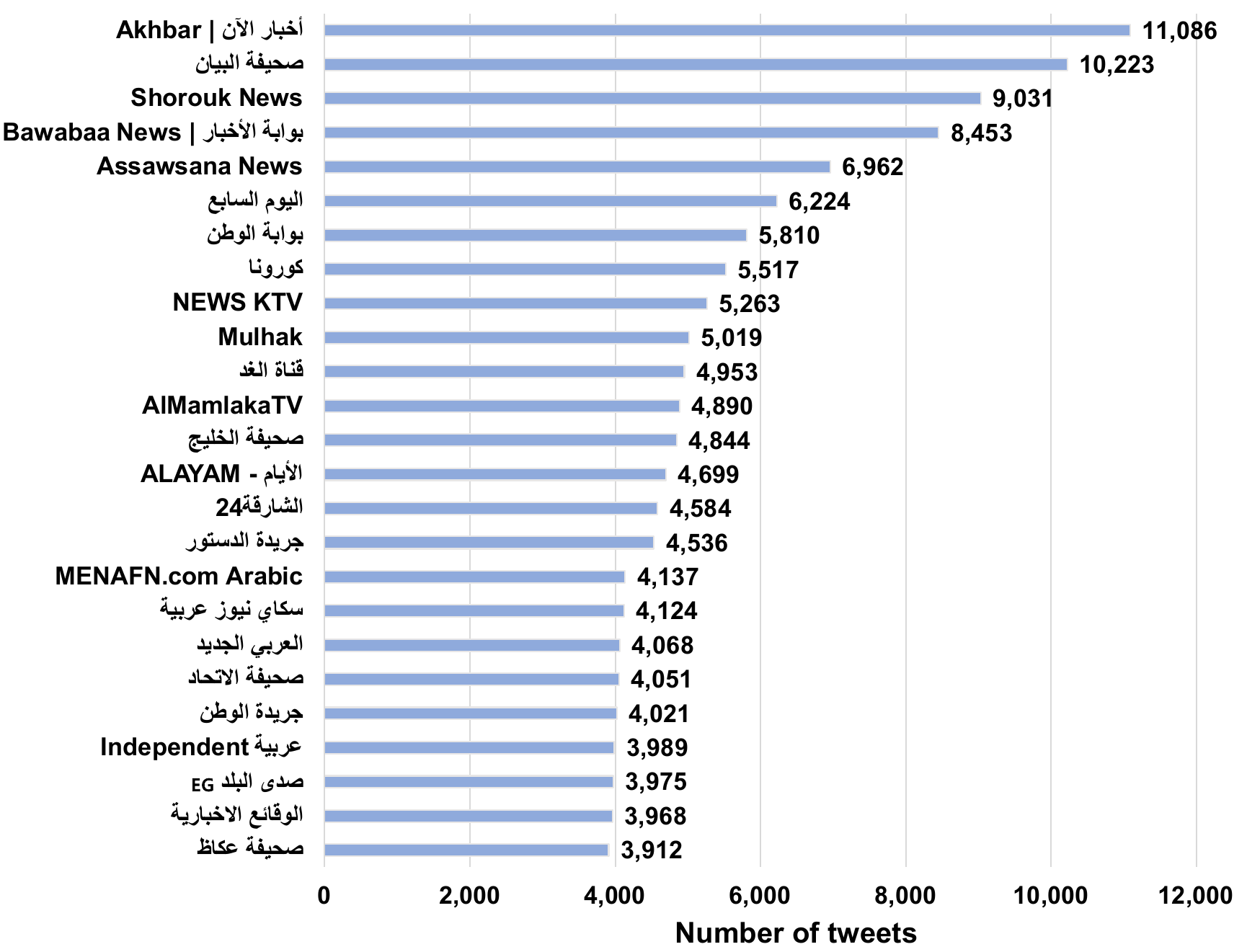}
  \caption{Number of tweets posted by the top 25 tweeters in \ds{} for the period of 27$^{th}$ Jan 2020 to 31$^{st}$ Jan 2021.\label{top25users}} 
\end{figure}

Figure~\ref{top25users} presents the top 25 tweeting users in \ds. We notice that several of them are news sources, and 20 are verified Twitter accounts. 

\begin{table}[h]
\centering
\caption{Tweets and users statistics of \ds. Stats for replies are for tweets until end of April 2020. 
\label{tab:statistics}}
  \resizebox{\columnwidth}{!}{%
    \begin{tabular}{lrl}
    \toprule
        \textbf{Tweets Statistics} && \\
    \midrule
        Source Tweets   & 2,675,049    \\ 
        Geotagged & 2,078 &(0.08\%)\\ 
        Geolocated & 60,873 &(2.28\%) \\
        Posted by verified users  & 499,351 &(18.66\%)   \\ 
        Include URL & 679,482 &(25.40\%)\\
        Include Media &  973,952 &(36.41\%)\\
        Top Subset & 370,132 &(13.84\%) \\ 
        Retweets of Top Subset & 7,925,821&\\
        Replies of Top Subset& 1,476,950&\\
    \midrule
        \textbf{Users Statistics} & \\
    \midrule
        Unique  & 690,339&  \\ 
        Verified & 5,575 &(0.81\%)     \\
        Average followers count & 4,630&   \\
        Average friends count & 918 & \\
        Average statuses count & 9,054&   \\
    \bottomrule
\end{tabular}%
}
\end{table}

\begin{figure*}[p]
\centering
  \includegraphics[scale=0.45]{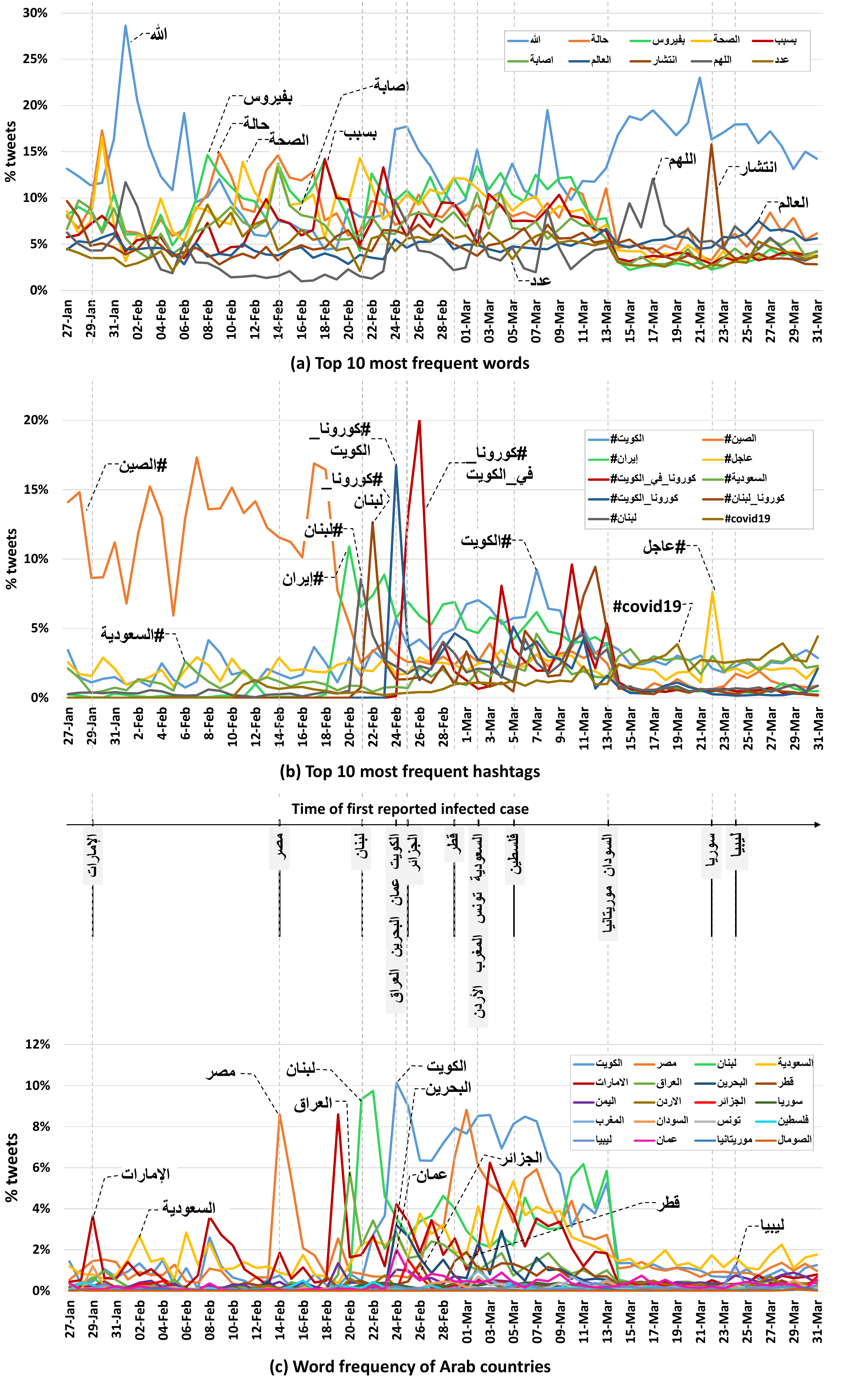}
  \caption{Time series of the frequency (in percentage of tweets) of top general keywords, hashtags, and Arab country words for the period of 27$^{th}$ Jan to 31$^{st}$ Mar. The time of first reported cases in Arab countries is also indicated and aligned with the time series.
  \label{fig:aligned}}
\end{figure*}

\subsection{Tweets Content \& Topics}
It is important to demonstrate that the tweets in \ds{} constitute a good representative sample of the Arabic tweets posted during the target period on COVID-19, and that they cover the prevalent topics discussed over Twitter during that period. 
To help examine this hypothesis, we analyzed the textual content of the tweets; in particular, we identified the most-frequent words, hashtags, and Arab country names for the tweets posted in the period of 27$^{th}$ Jan to 31$^{st}$ Mar 2020. We focus on this critical period since the virus started to spread in the Arab world during that time.
 We then tracked their frequency over time. Figure~\ref{fig:aligned} shows the time series (over days) of the three types of keywords.\footnote{When identifying the most frequent words and hashtags, we excluded the ones we used in our search queries since they are expected to be very frequent by definition.}

The 10 most-frequent words shown in Figure \ref{fig:aligned}(a) indicate two different types of words: those that are directly related to COVID-19 
(e.g., ``health'')
 and those that are not but related to prayers and supplications (e.g., 
``Allah/God'')). 
It is interesting to see the word
``Allah''
 is very frequent early on when the news about the virus started to spread (probably over discussions around whether the pandemic is a punishment from God or not, we believe), then declines over time only to become frequent again when the virus started to widely spread in the Arab world.

Figure \ref{fig:aligned}(b) demonstrates how 
``\#China'' 
hashtag was the most trending one from 27 of January until 20 of February, as COVID-19 was prevalent only in China and still not widely spread (at that time) in other countries. We can see how it started to be less trending as the number of cases started to decline in China by that date.\footnote{\china} On the other hand, when COVID-19 started to spread in the Arab world, other hashtags started to become viral. Furthermore, Figure \ref{fig:aligned}(b) shows that frequency spikes of 
``\#Iran'',
``\#Lebanon'', and
``\#Kuwait'' 
exactly match the confirmation dates of first reported cases in Iran,\footnote{\iran} Lebanon,\footnote{\lebanon} and Kuwait,\footnote{\kuwait} respectively.

To further analyze trending topics, Figure~\ref{fig:aligned} features the timeline of the first reported cases in the Arab countries and aligns them with the time series throughout the figure. 
Aligning the timeline with the series in Figure~\ref{fig:aligned}(c) reveals a significant match between the frequency peaks of several country names and the corresponding dates of first reported cases in those countries, most notably in UAE, Egypt, Lebanon, Kuwait, Oman, Bahrain, Algeria, and Libya. 

Figure~\ref{fig:aligned} also demonstrates the power of \ds{} in capturing further controversial and trending topics. Table~\ref{topics} shows a timeline covering dates of specific topics of discussion trending on social media around times of peaks in tweeting frequency in Figure~\ref{fig:aligned}(c).

\begin{table}[h]
\centering
\resizebox{\columnwidth}{!}{%
\begin{tabular}{|l|l|>{\raggedright}p{3cm}|l|}
\hline
{\bf Country} & {\bf Date} & {\bf Topic} & {\bf Related News} \\
\hline
UAE & Feb 19 & Yemeni Foreign Affairs Minister thanks UAE for evacuating Yemeni students from China & \url{http://tiny.cc/71qtmz} \\
\hline
Iraq & Feb 20 & Iraq announces closure of borders with Iran & \url{http://tiny.cc/yzrtmz}\\
\hline
\multicolumn{ 1}{|l|}{Egypt} &      Mar 2 & Egyptian health minister announces a visit to China & \url{http://tiny.cc/7iqtmz} \\
\cline{2-4}
\multicolumn{ 1}{|l|}{} & Mar 7 & Kuwait bans travels with Egypt including entry of Egyptian residents & \url{http://tiny.cc/lqstmz} \\
\hline
KSA & Mar 4-5 & Closure of The Grand Mosque in Mecca & \url{http://tiny.cc/iqrtmz} \\
\hline
\end{tabular}}
\caption{Examples of trending topics in social media matched by spikes in tweeting frequency in \ds\label{topics}}
\end{table}

\begin{figure}[t]
\centering
  \includegraphics[width=\columnwidth]{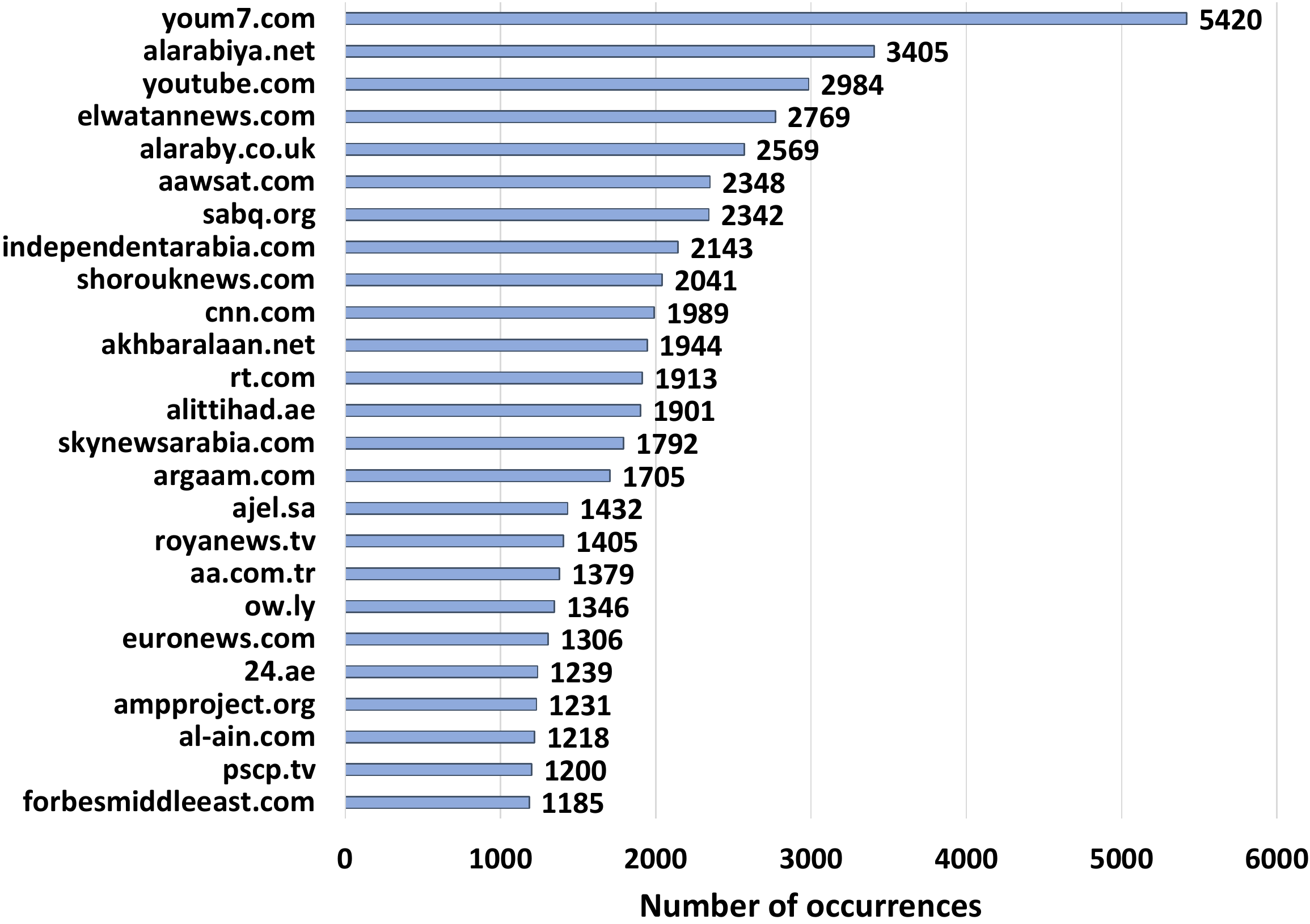}
  \caption{Most frequently-linked domains in the top tweets subset for the period of 27$^{th}$ Jan 2020 to 31$^{st}$ Jan 2021. 
\label{top25domains}}
\end{figure}

We further explore the topics discussed in \ds{} by considering the domains of URLs shared in the tweets. We focused on the top tweets subset of \ds{} (constructed as detailed in Section~\ref{propagationCollection}) and identified the URLs posted in those tweets. We then expanded them (since Twitter URLs are shortened) and dropped URLs of images and videos uploaded to Twitter, to focus only on URLs referring to external sources. We extracted 110,220 URLs from 2,617 unique domains. Figure~\ref{top25domains} shows that URLs from news websites are dominantly the most-commonly shared. We observe that these news websites mainly originate from three Arab countries, namely, Egypt, Saudi Arabia, and United Arab Emirates. Interestingly, videos from YouTube are among the most commonly shared media. Coupled with the fact that 36\% of the tweets in \ds{} include embedded images and videos (Table \ref{tab:statistics}), we believe this enables \ds{} to be a potential dataset to further support the evaluation of multi-modal retrieval and classification systems.


\subsection{Geographic Distribution of Tweets}
\label{sec:geo}
Although Twitter provides automatic geo-reference functionally, few users (solely around 1-3\% \cite{murdock2011your,jurgens2015geolocation}) opt to enable it due to privacy and safety reasons. Alternatively, to have an insight about the geographical distribution and diversity of the tweets in our dataset, we examined the \textit{place} and \textit{coordinates} attributes of the Tweet object.\footnote{\url{https://developer.twitter.com/en/docs/tweets/data-dictionary/overview/geo-objects}} We note that the \textit{place} attribute is an optional attribute that allows the user to select a location from a list provided by Twitter (therefore, the location might not necessarily show the actual location from where the tweet is posted), whereas the \textit{coordinates} attribute represents the geographic location of the tweet as reported by the user or client application.

Table~\ref{tab:statistics} shows that \ds{} has 
60,873 geolocated tweets (i.e., having values in the \textit{place} attribute) and 2,078 geotagged tweets (i.e., having values in the \textit{coordinates} attribute). Those tweets were posted by 24,072 and 256 unique users respectively. The geolocated tweets constitute about 2.28\% of the total source tweets, which is consistent with previous studies~\cite{huang2019large}.

\begin{figure}[h]
\centering
  \includegraphics[scale=0.27]{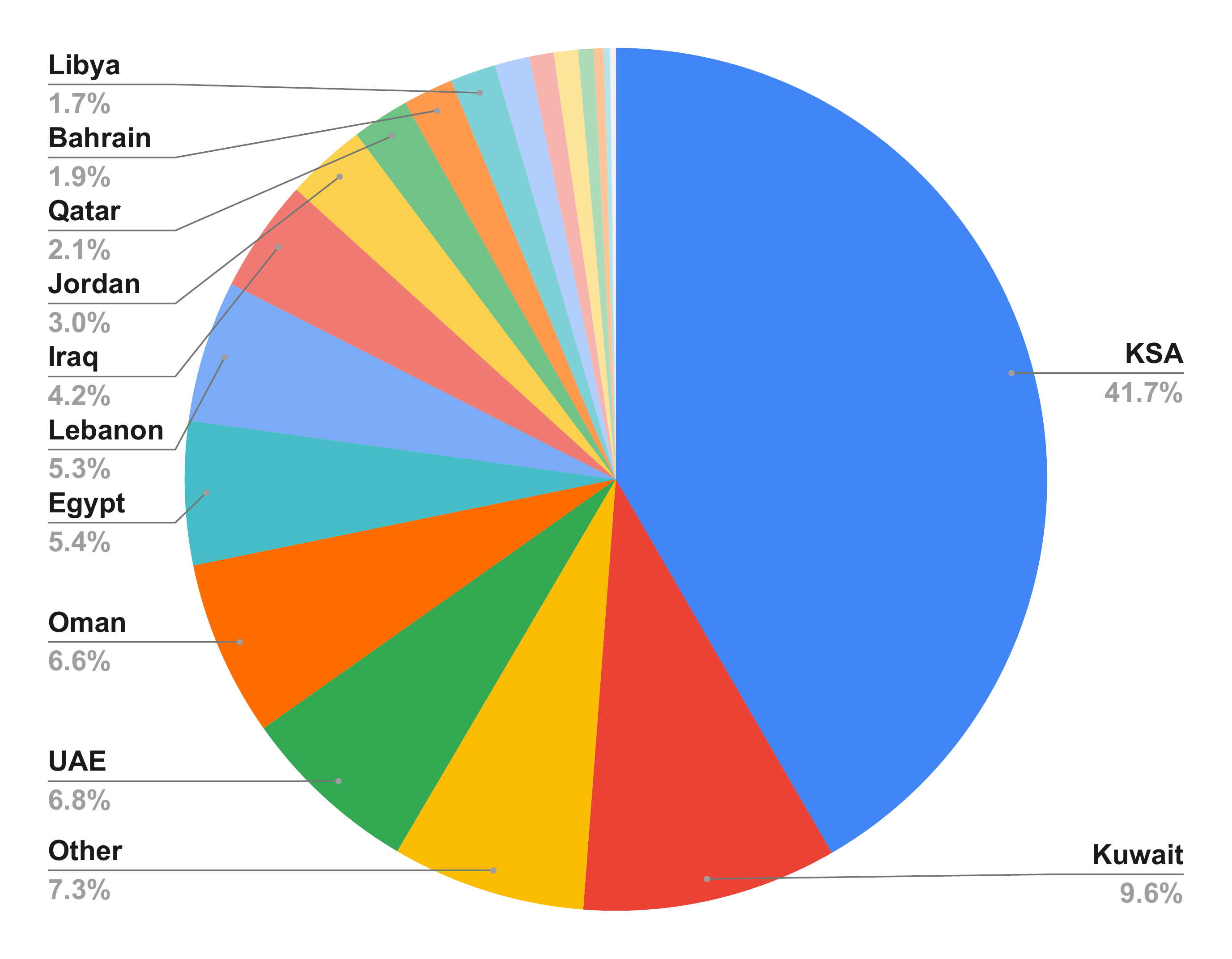}
  \caption{Distribution of geolocated tweets for the period of 27$^{th}$ Jan 2020 to 31$^{st}$ Jan 2021. \label{fig:countries_pie}}
\end{figure}

The geolocated tweets were indeed posted by users from 102 countries from around the globe. We found that 92.75\% of them were posted from the Arab world. The largest contribution of the geolocated content (about 41.7\%) comes from Saudi Arabia; this is somewhat expected as Saudi users represent the highest number of active Twitter users in the Arab world.\footnote{\twitterSaudi} Surprisingly, Kuwait comes second with 9.6\% of the geolocated content. We believe the rationale is that it was among the first countries that reported COVID-19 cases in the Middle East. Since then, Kuwait started a series of strict precautions such as a wide lock-down in many vital facilities until the government had imposed a nationwide curfew. We think, after the curfew, the people become more active on Twitter as a platform to break news and discuss developments of the virus. Additionally, we used related phrases and hashtags to ``\textit{Kuwait}'', ``\textit{Lebanon}'', and ``\textit{UAE}'' among the tracking keywords in the period between 22 of February and 13 of March. 
Furthermore, it is not surprising to see the countries that have a few 
cases have the smallest portions of contribution to the content. Others have small audience userbases on Twitter (e.g., Tunisia).


\subsection{Propagation Networks} 
\urldef{\retweetsAPI}\url{https://developer.twitter.com/en/docs/tweets/post-and-engage/api-reference/get-statuses-retweets-id}
As discussed earlier, we collected the propagation networks of the top subset. Overall, the number of retweets is 7,925,821 for the entire subset, and the number of replies is 1,476,950~\footnote{The stats are for the replies for data until end of April 2020 and we are still collecting the rest; we will make all available in the near future.}.

\begin{figure}
\centering
  \includegraphics[scale=0.23]{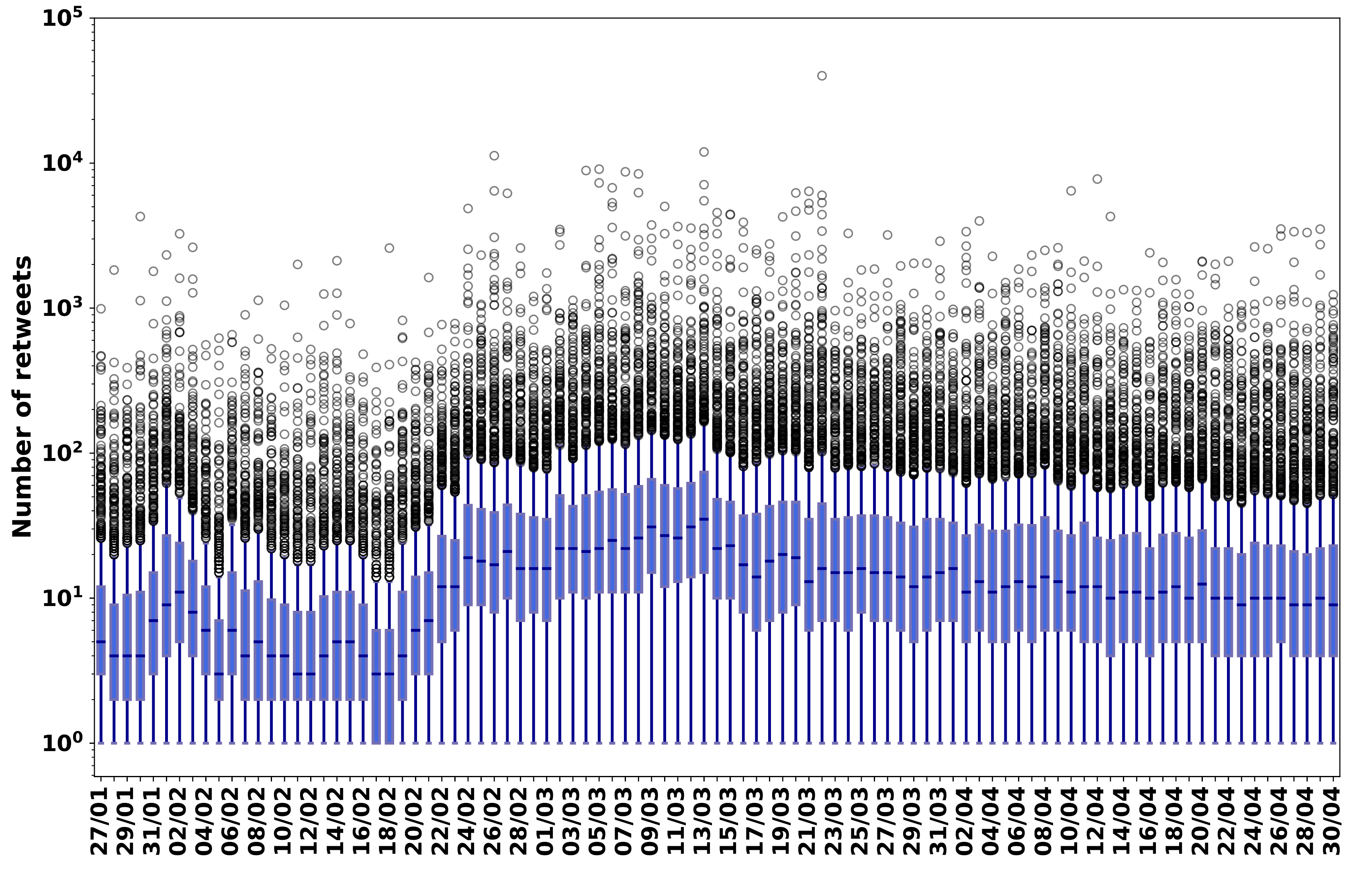}
  \caption{Distribution of retweets per day for the top subset for the period of 27$^{th}$ Jan 2020 to 30$^{th}$ April 2020.\label{fig:retweets}}
\end{figure}

\begin{figure}
\centering
  \includegraphics[scale=0.23]{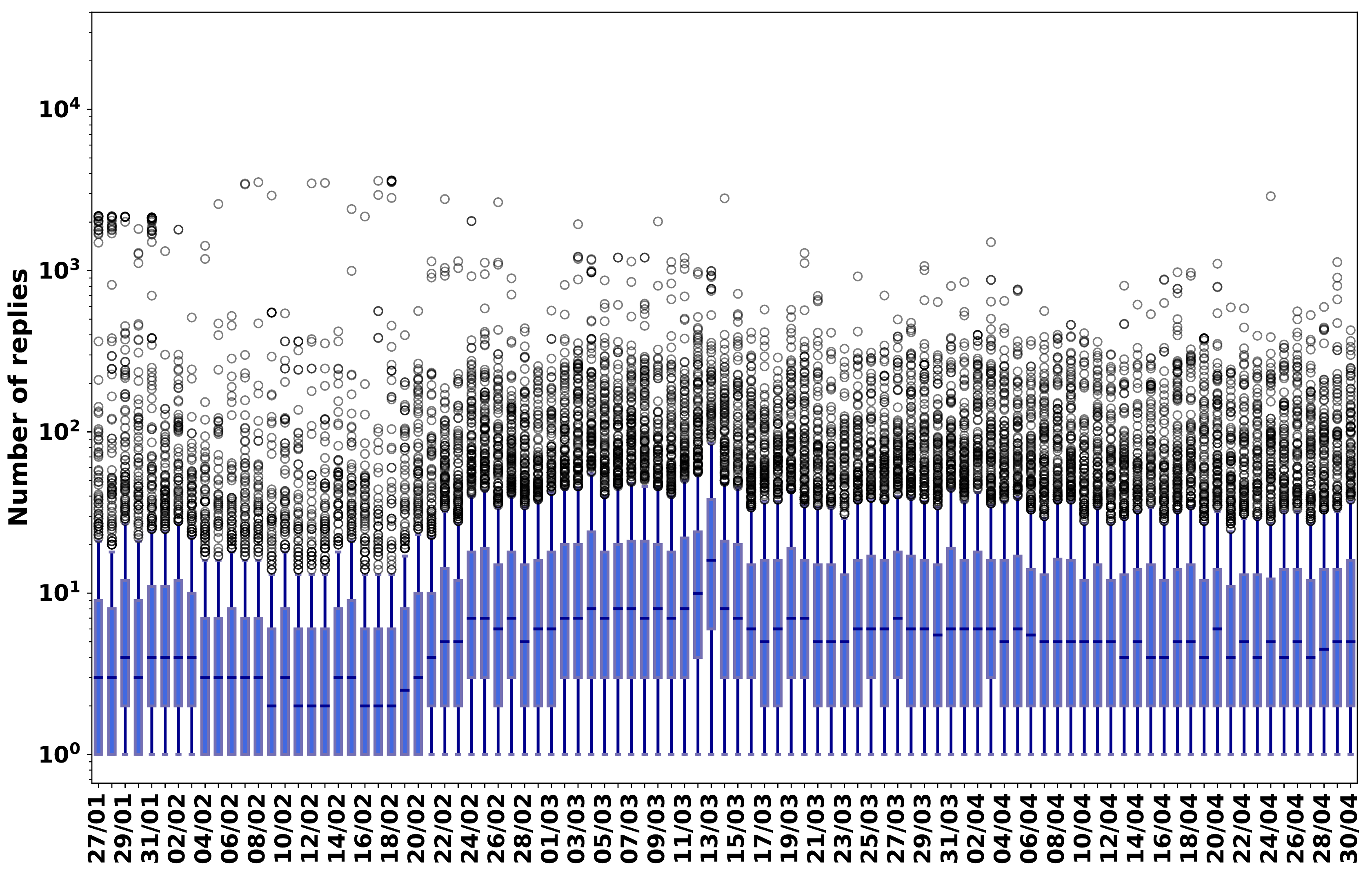}
  \caption{Distribution of replies per day for the top subset for the period of 27$^{th}$ Jan 2020 to 30$^{th}$ April 2020.}\label{fig:replies}
\end{figure}

At the time of collecting the retweets, we were not able to get the retweets of some of the tweets either because they were deleted or they were posted from private accounts. For those collected successfully, Figure~\ref{fig:retweets} illustrates the distribution of retweets per day using boxplots. It shows that the average (and median) number of retweets follows a similar pattern to what was shown in Figure~\ref{fig:ntweets} (notice that the Y-axis here is in log scale). We also notice that a good number of tweets got more than 100 retweets; some of them got even larger numbers reaching about 10k retweets or more, showing highly propagated content. As shown in Figure~\ref{fig:replies}, we applied a similar analysis to the replies per day, and found similar patterns to the retweets.

\section{Enabling Research} \label{usecases}
With the spread of Novel Coronavirus in several Arab countries and the subsequent procedural measures taken by the local governments, it continued to dominate discussions over social media (and Twitter in particular) to the time of this writing.
To enable research on different tasks on Arabic Tweets, we make \ds{} publicly-available for the research community. Having quantified the topical, geographical, and community
representativeness of \ds{}, we envision that it is suitable for diverse natural language processing, information retrieval, and computational social science research tasks including, but not limited to, emergency management, misinformation detection, and social analytics, as we discuss below. Furthermore, we provide propagation networks of daily highly popular subset (the most popular 1K) to ensure the quality of tweets. This sample is drawn after filtering out potential low-quality tweets (e.g., spam) and content duplicates. We anticipate the top subset and the network to support these tasks on the popular Arabic tweets that we assume have the most effect on shaping public opinion and understanding.

\textbf{Emergency Management}: As COVID-19 is an international pandemic, national and international health organizations need to analyze the effects of the outbreak. We believe \ds{} can support several tasks in that domain, such as filtering of informative content, events and sub-events detection, summarization, identification of eyewitnesses, geolocation, and studying information and situational awareness propagation, to name a few.

\textbf{Social Analytics}: In addition to sharing reports and awareness during the outbreak, people tend to discuss their opinions (e.g., stance towards the situation and its consequences) and express their emotions (e.g., big changes in lifestyle, social distancing, loss of their beloved ones, etc.). 
Therefore, analyzing the tweets to detect sentiment, stance, hate speech, and generally, offensive language, among other aspects, is of interest to many stakeholders. 

\textbf{Misinformation Detection}: With the sheer amount of information shared about COVID-19, many rumors are disseminated and getting high attention from the community, which causes a fast propagation of misinformation. This hinders the efforts of the health and governmental organizations on fighting the pandemic as such rumors spread panic and may lead to undesirable consequences (e.g., increase of cases and mortality rate, or lack of supplies due to hoarding). \ds{} supports studying information/claims propagation, claims check-worthiness detection and verification tasks on the most popular tweets. Furthermore, the retweet networks and conversational threads provide a valuable resource for early detection of fake news and identification of malicious rumor-spreading accounts. To further facilitate work in this domain of problems, we recently constructed ArCOV19-\emph{Rumors}, an annotated dataset on top of \ds{} to support claim and tweet verification ~\cite{haouari2020arcov19rumors}.

\section{Conclusion} \label{conc}
In this paper, we presented \ds{}, the first Arabic Twitter dataset about the Novel Coronavirus (COVID-19) 
that includes propagation networks of a large subset of tweets. We release all source tweets, top subset, search queries, and the propagation networks. Preliminary analysis showed that \ds{} captured spikes in tweeting frequency for country-specific tweets that are consistent with the first reported cases of COVID-19 in several Arab countries. We also found dominance of news agencies among top tweeting users and among most shared URLs. \ds{} enables research under many domains including natural language processing, information retrieval, and social computing. 
We plan to continue collecting tweets for the foreseeable future and the dataset will be continuously updated with newly collected tweets and propagation networks.
\section*{Acknowledgments}
The work of Tamer Elsayed and Maram Hasanain was made possible by NPRP grant\# NPRP 11S-1204-170060 from the Qatar National Research Fund (a member of Qatar Foundation). The work of Reem Suwaileh was supported by GSRA grant\# GSRA5-1-0527-18082 from the Qatar National Research Fund and the work of Fatima Haouari was supported by GSRA grant\# GSRA6-1-0611-19074 from the Qatar National Research Fund. The statements made herein are solely the responsibility of the authors. 
\bibliography{0.arcov19-main}
\bibliographystyle{acl_natbib}

\end{document}